\documentclass[man,floatsintext]{apa7}

\usepackage{amsmath,amssymb,amstext}
\usepackage{graphicx}
\usepackage{xcolor}
\usepackage{ulem} 

\usepackage{natbib}
\renewcommand{\cite}{\citep}

\AtBeginEnvironment{tabularx}{\tiny} 

\usepackage{booktabs}
\usepackage{tabularx}
\usepackage{ragged2e} 
\usepackage{rotating}
\usepackage{array} 
\usepackage{subcaption} 
\usepackage{tikz}

\usepackage{ifthen}

\title{{\bf Levels of Analysis for Large Language Models}}

\shorttitle{Levels of Analysis for Large Language Models}

\author{
Alexander Y. Ku$^{1}$,
Declan Campbell$^{2}$,
Xuechunzi Bai$^{3}$,
Jiayi Geng$^{4}$,
Ryan Liu$^{4}$,
Raja Marjieh$^{1}$,
R. Thomas McCoy$^{5}$,
Andrew Nam$^{6}$,
Ilia Sucholutsky$^{7}$,
Veniamin Veselovsky$^{4}$,
Liyi Zhang$^{4}$,
Jian-Qiao Zhu$^{4}$,
Thomas L. Griffiths$^{1,4,6}$
}

\authorsaffiliations{
$^{1}$\mbox{Department of Psychology, Princeton University},
$^{2}$\mbox{Princeton Neuroscience Institute, Princeton University},
$^{3}$\mbox{Department of Psychology, The University of Chicago},
$^{4}$\mbox{Department of Computer Science, Princeton University},
$^{5}$\mbox{Department of Linguistics, Yale University},
$^{6}$\mbox{Princeton Laboratory for Artificial Intelligence, Princeton University},
$^{7}$\mbox{Center for Data Science, New York University}
}

\abstract{%
Modern artificial  intelligence systems, such as large language models, are increasingly powerful but also increasingly hard to understand.
Recognizing this problem as analogous to the historical difficulties in understanding the human mind, we argue that methods developed in cognitive science can be useful for understanding large language models. We propose a framework for applying these methods based on the levels of analysis that David Marr proposed for studying information processing systems. By revisiting established cognitive science techniques relevant to each level and illustrating their potential to yield insights into the behavior and internal organization of large language models, we aim to provide a toolkit for making sense of these new kinds of minds.

}


\begin{document}

\maketitle


\section{Introduction}

The last decade has seen a series of breakthroughs in artificial intelligence (AI) research, culminating in the creation of the large language models that underlie chat-based agents such as ChatGPT, Claude, Gemini, and LLaMa \cite{achiam2023gpt,team2023gemini,touvron2023llama}. These breakthroughs have been driven by a specific strategy: starting with generic artificial neural network architectures and increasing their size and training data. Artificial neural networks are notoriously difficult to interpret, finding solutions to problems that are expressed in the form of billions of continuous weighted connections between units. As a consequence, computer scientists now face an unfamiliar problem: they have created systems that they do not understand. To make things even worse, since the training data and weights of many of the leading systems are not available outside the companies that created them, in many cases the only insights we can obtain about the nature of these systems are those that can be gleaned by studying their behavior.

Even though this problem is unfamiliar to computer scientists, it is very familiar to another group of researchers: cognitive scientists. Cognitive science is the interdisciplinary science of the mind, and for the 70 or so years since its inception \cite{miller03} has been limited by the fact that it had relatively few kinds of mind to study. To cognitive scientists, the advent of intelligent machines offers exciting new opportunities to apply methods that have been refined through trying to understand how human minds work \cite{binz2023using,coda2024cogbench}. Those methods encompass both techniques that are based on human behavior and insights that come from related fields such as neuroscience that explore the underlying mechanisms.

One powerful conceptual framework used in cognitive science is the idea that information processing systems can be understood at different levels of analysis. The computational neuroscientist David Marr proposed three such levels \cite{marr82}: the {\em computational} level, which focuses on the abstract computational problem a system solves and its ideal solution; the {\em algorithmic} level, which focuses on the representations and algorithms that approximate that solution; and the {\em implementation} level, which focuses on how those representations and algorithms are realized in a physical system. The same three levels can be used for analyzing large language models, focusing on how such systems are shaped by their function, the solutions that they seem to find, and the realization of those solutions in weights and units within the underlying artificial neural network (see Table \ref{tab:marr_analogy}). 

Drawing these parallels between the levels at which we analyze minds and machines provides a guide for where we might expect to look for relevant tools for understanding the behavior of AI systems: analyses at the computational level can make use of computational modeling techniques, analyses at the algorithmic level can draw on methods from cognitive psychology, and analyses at the implementation level can take inspiration by neuroscience. In this paper we highlight some of the tools from cognitive science that we believe are particularly useful for understanding large language models (and related approaches such as vision-language models) at each of these levels -- techniques such as rational analysis, the axiomatic approach, and multidimensional scaling. We illustrate the potential of this approach by using examples from our own work and by drawing analogies between nascent methods in computer science and tools developed in other fields.

\begin{table}[t!]
  \centering
  \caption{Understanding natural and artificial minds across Marr's levels of analysis.}
  \label{tab:marr_analogy}
  \begin{tabularx}{\textwidth}{>{\RaggedRight}p{0.2\textwidth} >{\RaggedRight}p{0.2\textwidth} >{\RaggedRight}X >{\RaggedRight\arraybackslash}X}
    \toprule
    \textbf{Level of analysis} & \textbf{Inquiry} & \textbf{Cognitive science} & \textbf{Artificial intelligence} \\
    \midrule
    \textbf{Computational} & What problem is being solved and what is the ideal solution? & Understanding behavior in terms of optimal solutions to environmental pressures (e.g., Bayesian models of cognition). & Understanding behavior in terms of the training objective (e.g., next-token prediction) and deviations from optimal benchmarks (e.g., violations of probability axioms). \\
    \addlinespace
    \textbf{Algorithmic} & How is the solution approximated and what representations and processes are involved? & Inferring mental representations and processes through behavioral experiments (e.g., reaction times, error patterns, similarity judgments). & Inferring representations and processes via behavioral analysis (e.g., analyzing systematic errors, soliciting similarity judgments, uncovering hidden associations). \\
    \addlinespace
    \textbf{Implementation} & How are the representations and processes physically realized? & Studying neural circuits and population activity using techniques like optogenetics, single-unit recording, fMRI, and MVPA. & Studying artificial neurons, circuits (e.g., induction heads), and population activity using techniques like activation patching, sparse autoencoders, and representational geometry analysis. \\
    \bottomrule
  \end{tabularx}
\end{table}

\section{Computational Level}

The computational level focuses on the abstract problem that a system needs to solve. By recognizing the pressures that this problem exerts upon the system, we can make predictions about the properties that the system is likely to have \cite{shepard87,marr82,anderson1990adaptive,griffiths2020understanding}. This perspective is perhaps most familiar from evolutionary biology, in which organisms are understood through the lens of the evolutionary pressures that have shaped them. For instance, our understanding of bird flight is informed by the aerodynamic principles that bird flight must obey. Similarly, we can gain insight into intelligent systems by considering how they have been shaped by the functions that they must perform.

This emphasis on function makes the computational level well-suited for analyzing AI systems. Although many aspects of modern AI systems are difficult to interpret (including their behavior and the mechanisms that they use to achieve that behavior), one aspect that we understand well is the function that the system is optimized to perform. Specifically, this function is explicitly defined by humans in the form of the AI system’s training objective. Therefore, computational-level analysis enables us to start from something we understand well -- the training objective -- and use it to reason about the less-well-understood territory of the behavior of the resulting systems.

\subsection{Embers of Autoregression}

\citet{mccoy2023embers} used an approach based on computational-level analysis to try to understand the behavior of large language models (LLMs). For these systems, the primary training objective is next-token prediction, also known as autoregression: predicting the next token (word or part of a word) in a piece of text given the preceding tokens. Analysis of this task leads to the prediction that LLMs will perform better when they need to produce a high-probability piece of text than when they need to produce a low-probability piece of text, even in deterministic settings where probability should not matter. Intuitively, this prediction follows from the way in which next-token prediction fundamentally depends on the probabilities of token sequences; this intuition is derived more formally in \citet{mccoy2023embers} via a Bayesian analysis of autoregression.

\begin{figure}[t!]
    \centering
    \includegraphics[width=\linewidth]{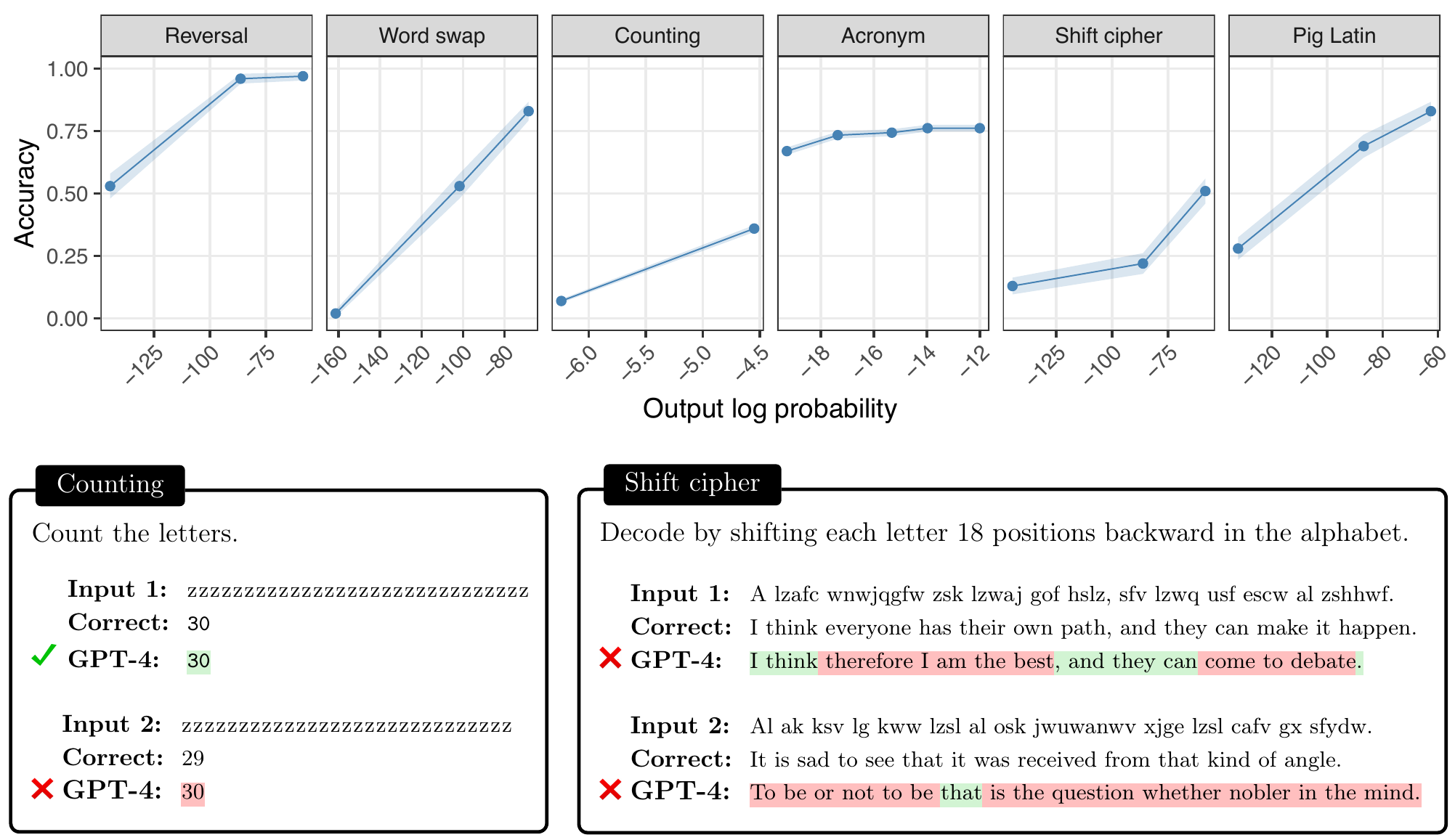}
    \vspace{2mm}
    \caption{Large language models perform better when they need to produce a high-probability piece of text than when they need to produce a low-probability piece of text, even in deterministic settings where probability should not matter.}
    \label{fig:embers}
\end{figure}

The prediction that LLM behavior will be sensitive to probability is borne out in experiments testing a range of LLMs across a range of tasks (see Figure \ref{fig:embers}). For instance, when GPT-4 is asked to count how many letters there are in a list, it performs much better when the answer is a frequently-used number than a more rarely-used number; e.g., when the answer is 30, its accuracy is 97\%, but when the answer is 29, its accuracy is only 17\%, presumably because the number 30 is used more often in natural text than the number 29. Similarly, when asked to decode a message written in a simple cipher, GPT-4’s accuracy was 51\% when the answer was a high-probability sentence but only 13\% when the answer was a low-probability word sequence. Thus, even though LLMs can be applied to many different tasks -- a capability that has been viewed as evidence that LLMs show ``sparks of artificial general intelligence'' \cite{bubeck2023sparks} -- they also continue to display ``embers of autoregression'' -- behavioral trends that reflect the nature of the specific objective they were optimized for.

\subsection{Bayesian Optimality as a Benchmark}

While the ``embers of autoregression'' reflect the fact that large language models are pretrained on text prediction, the use of this objective may also mean that we can understand their behavior in terms of Bayesian inference more generally. Accurately predicting the next token in text requires making implicit inferences  not just about what sequences of words are related to one another, but about events in the world described by those words. Bayesian models of cognition provide optimal solutions to such problems, and have become instrumental in explaining human performance across diverse domains, including perception \cite{yuillek06}, language processing \cite{chaterm06, griffithsst07}, categorization \cite{sanborngn10}, and intuitive physics \cite{sanbornmg13, Battaglia2013}. The emergence of a broader capability to draw inductive inferences about the world suggests that we can likewise compare LLMs to these optimal solutions to gain insight into their behavior, much as cognitive scientists use Bayesian models to understand human cognition.

The connection between Bayesian optimality and LLMs can be made more explicit by considering the problem of next-token prediction that is typically used in training these models. Predicting the next token can be done by extracting the predictive sufficient statistics from previous tokens \cite{bernardos94}. For some datasets, the Bayesian posterior distribution over particular parameters or hypotheses about the generating process can serve as a predictive sufficient statistic \cite{zhang2024embeddingsembedautoregressivemodels}. This perspective can be used to understand the representations that LLMs form and how they should relate to ideal Bayesian solutions \cite{zhang2023deepfinettirecoveringtopic,
zhang2024embeddingsembedautoregressivemodels}. Several other recent papers have also identified interesting connections between LLMs and Bayesian inference \cite{xie2021explanation,wang2023large,zheng2023revisiting}.

These connections suggest that we might be able to create simple Bayesian models of the inferences drawn by LLMs. Such Bayesian models can be used to explore the implicit prior distributions adopted by LLMs and to compare the resulting distributions with those inferred from human behavior. For example, \citet{griffiths2024bayes} used a simple Bayesian model of predicting the future \cite{griffithst06future} to recover implicit prior distributions about the extent or duration of phenomena from GPT-4. \citet{zhu2024eliciting} built on this work, using an iterated learning procedure originally developed for sampling from human priors to explore the priors that LLMs have for causal relationships and probability distributions, as well as their implicit assumptions about speculative events such as the development of superhuman AI. Bayesian analyses can also prove fruitful in understanding more complex aspects of the behavior of AI systems, such as when methods like chain of thought prompting are effective \cite{prystawski2023think}
and when models will engage in in-context learning \cite{wurgaft2025context}.

Importantly, Bayesian models can be useful for understanding a system even if that system only roughly approximates Bayesian inference. Even in such cases, knowing the ideal solution to the problem that the system is solving can be helpful in making sense of its behavior. In particular, knowing where the system deviates from a Bayesian model can be valuable. These deviations can be used to identify behavior that needs to be explained at other levels of analysis. For example, the approach of resource-rational analysis \cite{lieder2020resource} is based on identifying possible algorithmic-level mechanisms based on deviations from ideal solutions.

\subsection{Violations of Axiomatic Systems}

Bayesian inference characterizes the optimal solution to a particular computational problem. Comparisons to a Bayesian model are typically done in a quantitative way, directly assessing how well the model captures the behavior. Another approach to understanding intelligent systems at the computational level is to provide a qualitative comparison of their behavior to a formal axiomatic system that describes the ideal solution to a problem. In this type of analysis, the goal is not to model the system's inductive inferences, but to test its coherence against the logical rules of a domain, such as arithmetic or probability theory.

One of the original examples used by Marr had this flavor: he suggested that we can understand the computational problem solved by a cash register by recognizing that the expectations we have about shopping, such as the fact that the order in which items are checked out doesn't affect the total price, correspond to the axiomatic system of arithmetic. In cognitive science, the most celebrated application of this approach has been decision theory. By considering how to define rationality, decision theorists were able to specify a set of axioms that result in the discovery that rational agents should seek to maximize expected utility \cite{vonneumann47,savage54}. Asking whether this axiomatic system actually describes human behavior resulted in fundamental insights into human decision-making, with Kahneman and Tversky carrying out an influential research program that showed that people systemically violate the prescriptions of these axioms \cite{tverskyk74,kahneman1979prospect}.

\begin{figure}[t!]
    \centering
    \begin{tikzpicture}
        \node[inner sep=0pt, anchor=south west] (image) at (0,0) {\includegraphics[width=\linewidth]{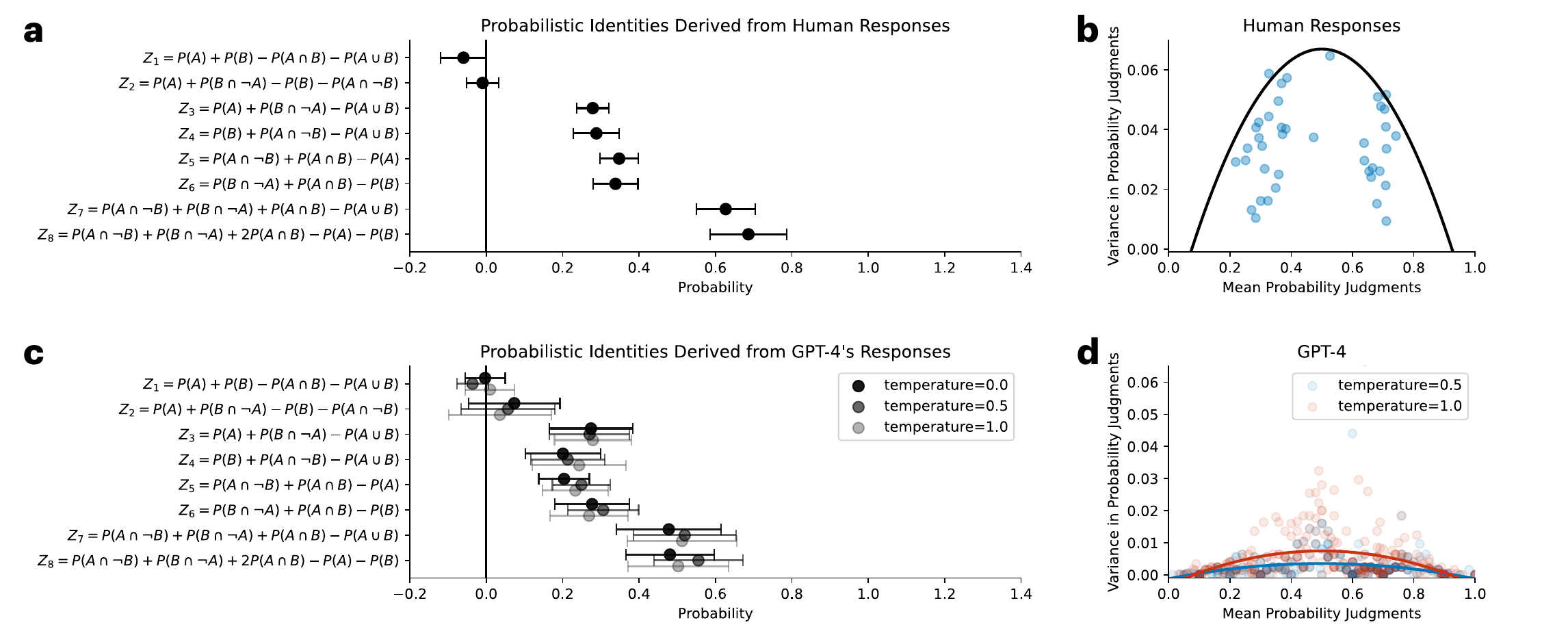}};
        
        \begin{scope}[x={(image.south east)},y={(image.north west)}]
            
            \node[fill=white, minimum width=16pt, minimum height=16pt, anchor=north west, xshift=2pt, yshift=-2pt] at (0.0, 1.0) {};
            \node[anchor=north west, xshift=-2pt, yshift=2pt, font=\large\sffamily] at (0.0, 1.0) {\textbf{a}};
            
            \node[fill=white, minimum width=16pt, minimum height=16pt, anchor=north west, xshift=2pt, yshift=-2pt] at (0.67, 1.0) {};
            \node[anchor=north west, xshift=-2pt, yshift=2pt, font=\large\sffamily] at (0.66, 1.0) {\textbf{b}};
            
            \node[fill=white, minimum width=16pt, minimum height=16pt, anchor=north west, xshift=2pt, yshift=-2pt] at (0.0, 0.5) {};
            \node[anchor=north west, xshift=-2pt, yshift=2pt, font=\large\sffamily] at (0.0, 0.5) {\textbf{c}};
            
            \node[fill=white, minimum width=16pt, minimum height=16pt, anchor=north west, xshift=2pt, yshift=-2pt] at (0.67, 0.5) {};
            \node[anchor=north west, xshift=-2pt, yshift=2pt, font=\large\sffamily] at (0.66, 0.5) {\textbf{d}};
            
        \end{scope}
    \end{tikzpicture}
    
    \caption{Incoherent probability judgments from humans (\textbf{a}, \textbf{b}) and GPT-4 (\textbf{c}, \textbf{d}). Like human probability judgments (\textbf{a}), GPT-4’s judgments systematically deviate from zero when combined into probabilistic identities (\textbf{c}). When repeatedly queried about the same event, the mean-variance relationship of probability judgments follows an inverted-U shape for both humans (\textbf{b}) and GPT-4 (\textbf{d}). Human data are adapted from \cite{zhu2020bayesian}, GPT-4 results are from \cite{zhu2024incoherent}.}
    \label{fig:violation_of_prob_theory}
\end{figure}

Considering relevant axiomatic systems -- and discovering how they are violated -- provides another tool for understanding LLMs.
For example, probability theory dictates that the probabilities of an event $A$ and its complement $\neg A$ sum to 1, meaning $P(A)+P(\neg A)=1$. To assess whether LLMs adhere to this rule, we can examine deviations from zero in $P(A)+P(\neg A)-1$. Similarly, other probabilistic identities can be tested, such as $P(A)+P(B)-P(A \land B)-P(A \lor B)$, which should also equal zero if judgments are coherent \cite{zhu2020bayesian}.
Eliciting probability judgments for logically related events from GPT-4 (see Figure \ref{fig:violation_of_prob_theory}) shows that probability identities formed using judgments generated by the LLM systematically deviate from zero, violating the rules of probability theory \cite{zhu2024incoherent}. In addition to maintaining coherence across logically related events, a rational agent should produce consistent probability judgments when repeatedly queried about the same event. However, repeated probability judgments elicited from LLMs exhibit an inverted-U-shaped mean-variance relationship \cite{zhu2024incoherent}. These systematic deviations also qualitatively mirror those observed in human probability judgments, suggesting that LLMs exhibit similar biases in probabilistic reasoning.

Optimal behavior can also be defined with respect to problems or situations with inherent uncertainty. For example, in a risky choice task where participants select one of many gambles (each gamble corresponding to a probabilistic distribution of outcomes), there is always a rational choice that maximizes the expected value (making the simplest possible assumption about the utility associated with an option by equating its utility with its monetary value). Here, using chain-of-thought reasoning \cite{wei2022chain} make choices almost completely rationally, but without such reasoning, their choices are noisy and sometimes ignore probabilities completely \cite{liu2024largelanguagemodelsassume}. Furthermore, when LLMs are asked to predict human performance on the task, they predict humans to behave highly rationally, even though people behave much less so.

Stepping slightly further afield, axiomatic analyses have also been applied in the domain of reasoning. For example, a classic finding in psychological studies of deductive reasoning is that people demonstrate ``content effects,'' being more likely to judge logical arguments to be valid when they agree with the conclusion \cite{evans1983conflict}. This is a violation of the properties of the axiomatic system of deductive logic, where the validity of an argument does not depend on its content. Interestingly, large language models demonstrate the same kind of bias \cite{lampinen2024language}. 

The variable performance of LLMs against normative benchmarks highlights the classic cognitive science distinction between competence (the underlying ability) and performance (the observed behavior). A model’s failure to behave rationally in a specific context may not indicate a fundamental lack of competence, but rather reflect how performance is shaped by factors like the query’s phrasing or the evaluation method itself. Therefore, assessing the true capabilities of LLMs requires careful consideration of the conditions under which those capabilities are measured \cite{lampinen2024can,hu2024auxiliary}.

\subsection{Summary} 

The computational level of analysis provides a powerful lens for understanding large language models by focusing on their function. Examining the training objective (in this case, autoregression) can directly predict specific behavioral patterns, as illustrated by the ``embers of autoregression.'' Furthermore, comparing LLMs to optimal benchmarks, whether derived from Bayesian models of cognition or the axioms of probability and decision theory, can reveal both the surprising capabilities and the systematic limitations of these systems. By considering what computational problem LLMs are solving (or approximating), we can gain significant insight into their behavior and internal structure, even when the algorithmic and implementation details remain opaque. 

\section{Algorithmic Level}

Just as the computational level asks what problem an information-processing system solves, the algorithmic level explores how that solution is approximated. This level concerns itself with the specific representations and algorithms used to carry out the computation. Consider bird flight again: while aerodynamics dictates the principles of flight (lift, drag, thrust), different bird species employ different algorithms – variations in flapping patterns and soaring techniques – to achieve flight. These variations represent different algorithmic solutions to the same computational problem. In cognitive science, understanding the algorithmic level involves designing experiments that probe the internal workings of the mind, inferring the nature of mental representations and the processes that operate on them. Just as ornithologists might study wing movements and muscle activity to understand bird flight, cognitive psychologists use reaction times, error patterns, and carefully designed stimuli to understand human cognition. We can apply similar techniques to investigate the algorithmic solutions of large language models.

The algorithmic level is particularly relevant to LLMs because, unlike traditional symbolic AI systems with explicitly programmed rules, the specific algorithms and representations employed by LLMs are not pre-defined. They are learned through the training process, resulting in complex and often opaque internal structures that aren't obviously localized in any particular location in the network. Proprietary closed-source networks present additional challenges as their internal states are not directly observable. This problem is much the same as that faced by psychologists before the advent of modern neuroscientific tools and methodologies.

Cognitive psychology offers a rich toolbox of methods for exploring the algorithmic level, many of which can be creatively adapted to study LLMs.
This section  explores how cognitive science-inspired approaches -- such as analyzing systematic error patterns, soliciting similarity judgments, and exploring associations -- can be used to uncover the algorithms and representations of LLMs.

\subsection{Parallel and Serial Processing}

A fundamental challenge for any cognitive system, whether biological or artificial, is the tradeoff between processing information serially (one item at a time) or in parallel (multiple items simultaneously) \cite{treisman1980feature,townsend1990serial}. Parallel processing offers efficiency, allowing for rapid processing of multiple inputs. However, it also introduces the potential for representational interference, especially when dealing with compositional representations, where features are shared and recombined across different items. Think of trying to remember a set of colored shapes: if the colors and shapes are reused across multiple objects, it becomes harder to keep track of which color goes with which shape when processing them all at once. Serial processing, while slower, mitigates this interference by focusing attention on a single item at a time.

\begin{figure}[t!]
    \centering 
    
    \begin{minipage}[c]{0.26\textwidth} 
        \begin{tikzpicture}
            \node[inner sep=0pt, anchor=south west] (image) at (0,0) {\includegraphics[width=\linewidth]{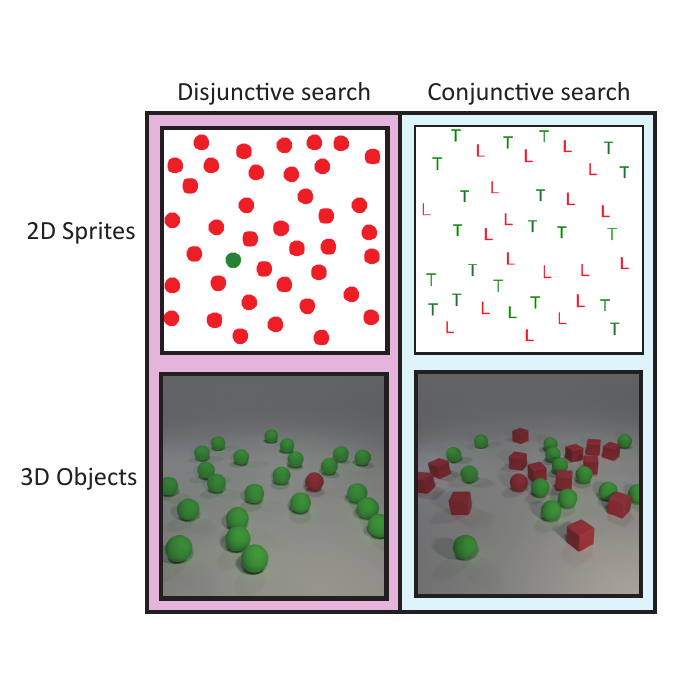}};
            \node[anchor=north west, xshift=-2pt, yshift=2pt, font=\large\sffamily] at (image.north west) {\textbf{a}};
        \end{tikzpicture}
    \end{minipage}\hspace{0.01\textwidth}
    \begin{minipage}[c]{0.19\textwidth} 
        \includegraphics[width=\linewidth]{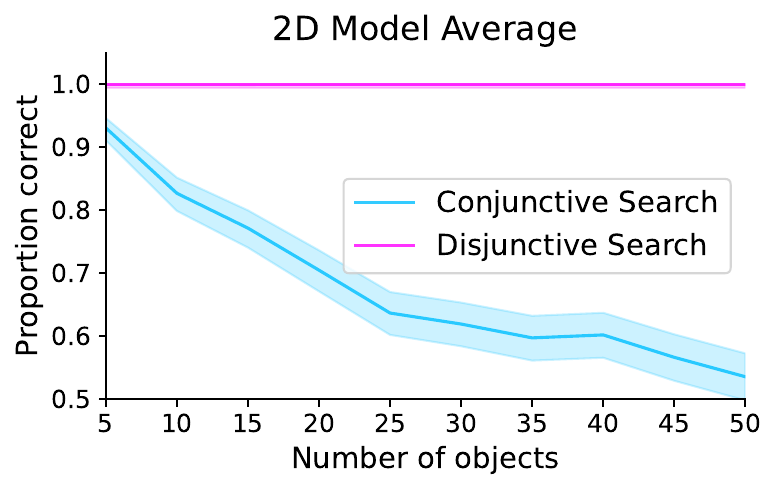} \\[4pt]
        \includegraphics[width=\linewidth]{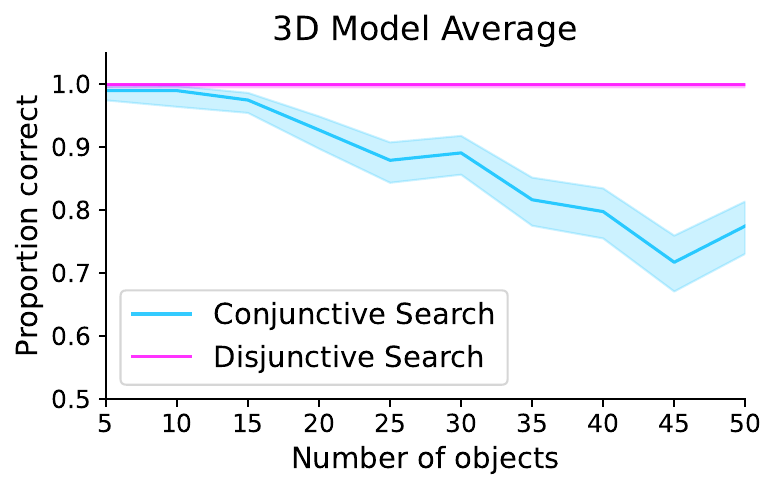}
    \end{minipage}\hfill 
    \begin{minipage}[c]{0.25\textwidth} 
        \begin{tikzpicture}
            \node[inner sep=0pt, anchor=south west] (image) at (0,0) {\includegraphics[width=\linewidth]{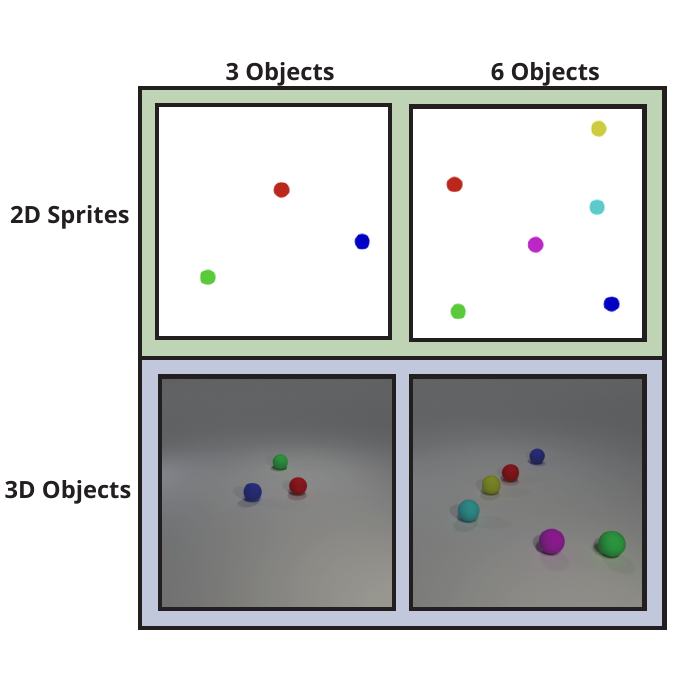}};
            \node[anchor=north west, xshift=-2pt, yshift=2pt, font=\large\sffamily] at (image.north west) {\textbf{b}};
        \end{tikzpicture}
    \end{minipage}\hspace{0.01\textwidth}
    \begin{minipage}[c]{0.26\textwidth} 
        \includegraphics[width=\linewidth]{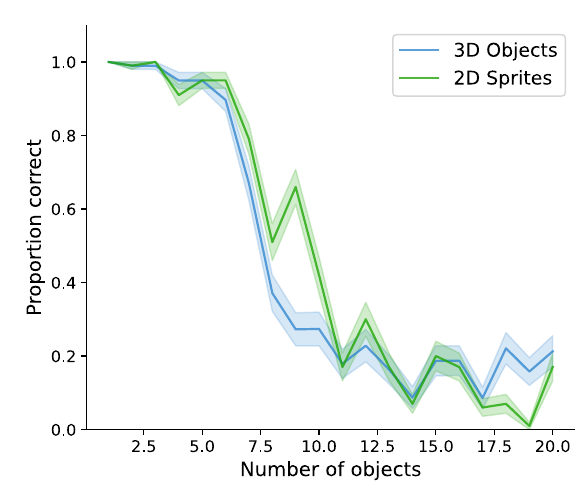}
    \end{minipage}
    
    \caption{Patterns of behavior consistent with parallel processing in vision-language models (VLMs). (\textbf{a}) VLMs show highly accurate ``pop-out'' search for distinctive visual targets but exhibit degraded performance in conjunction search as the number of distractors increases. (\textbf{b}) VLMs exhibit a ``subitizing limit'' in numerical estimation.}
    \label{fig:parallel}
\end{figure}

Critically, the interference caused by parallel processing of compositional representations follows a predictable pattern: items that share more features will interfere with each other more than dissimilar items \cite{musslick2021rationalizing,bouchacourt2019flexible}. This relationship between feature similarity and performance degradation provides a diagnostic tool. By observing systematic errors -- such as decreased accuracy or the formation of ``illusory conjunctions'' (e.g., misremembering a red square and a blue circle as a red circle) – we can infer that the system likely relies on compositional representations and parallel processing. This approach allows us to indirectly examine the structure of a system's representations by analyzing its behavioral limitations.

Recent work with vision-language models (VLMs) provides compelling evidence for this approach. VLMs are typically built on top of an LLM backbone, adding a system for encoding visual images and additional training on tasks that involve both language and images \cite{liu2023visual,bordes2024introduction}. The resulting models, like humans, show highly accurate ``pop-out'' search for distinctive visual targets but exhibit degraded performance in conjunction search (searching for a target defined by a combination of features) as the number of distractors increases \cite{campbell2024understanding}. This pattern suggests interference arising from the simultaneous processing of multiple items. Similarly, VLMs exhibit a ``subitizing limit'' in numerical estimation (see Figure \ref{fig:parallel}), akin to that observed in humans under conditions that force rapid, parallel visual processing \cite{kaufman1949discrimination, trick1994small, rane2024can}. Furthermore, when tasked with describing the features of multiple objects in a scene, VLMs make systematic errors resembling illusory conjunctions observed in human visual working memory tasks \cite{treisman1982illusory}, with error rates predicted by the potential for feature interference \cite{campbell2024understanding}. These parallels suggest that both humans and VLMs rely on compositional representations and are susceptible to similar forms of interference during parallel processing.

\subsection{Similarity Judgments}

Building on the principle that behavioral limitations can reveal representational structure, we now turn to specific methods for probing these representations in LLMs. One powerful approach, adapted from cognitive psychology, is the use of similarity judgments. Humans rely on efficient representations to navigate high-dimensional environments and to support different cognitive capacities \cite{anderson1990adaptive}. Characterizing the structure of those representations has been central to decades of psychological research spanning a wide array of contexts, including sensory domains such as color \cite{shepard80,ekman1954dimensions}, pitch \cite{shepard1982geometrical}, and natural images \cite{hebart2020revealing,marjieh2024universal}, linguistic domains such as the semantic organization of concepts \cite{rosch1975cognitive,tverskyh86} and lexical analogies \cite{peterson2020parallelograms}, and numerical domains such as the relations between integers and their mathematical properties \cite{miller1983child,pitt2022exact,piantadosi2016rational,tenenbaum1999rules}.

\begin{figure}[t!]
    \centering 
    
    \begin{minipage}[c]{0.49\textwidth}
        \begin{tikzpicture}
            \node[inner sep=0pt, anchor=south west] (image) at (0,0) {\includegraphics[trim={1.3in 2.23in 1.2in 0.0in}, clip, width=\linewidth]{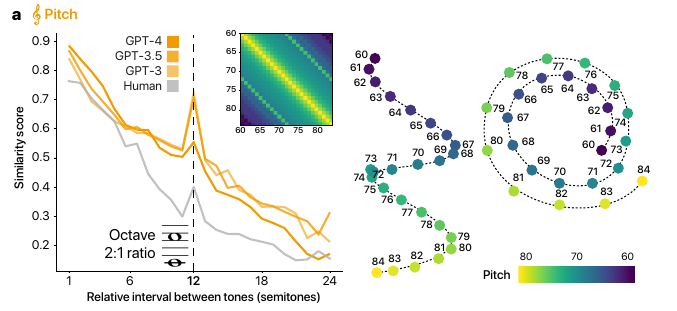}};
            
            \node[fill=white, minimum width=8pt, minimum height=8pt, anchor=north west] at (image.north west) {};
            
            \node[anchor=north west, xshift=-8pt, yshift=4pt, font=\large\sffamily] at (image.north west) {\textbf{a}};
        \end{tikzpicture}
    \end{minipage}\hfill
    \begin{minipage}[c]{0.49\textwidth} 
        \begin{tikzpicture}
            \node[inner sep=0pt, anchor=south west] (image) at (0,0) {\includegraphics[trim={1.3in 0in 1.2in 2.07in}, clip, width=\linewidth]{figures/similarity.pdf}};
            
            \node[fill=white, minimum width=8pt, minimum height=8pt, anchor=north west] at (image.north west) {};
            \node[anchor=north west, xshift=-8pt, yshift=4pt, font=\large\sffamily] at (image.north west) {\textbf{b}};
        \end{tikzpicture}
    \end{minipage}
    
    \caption{Exploring the sensory representations of large language models with similarity judgments. (\textbf{a}) For musical pitch, both humans and LLMs show a decrease in judged similarity with increases in the interval between tones, but also show an increase at tones an octave apart (a full similarity matrix for GPT-3 is shown inset). As a consequence, both human and LLM similarities are best captured by helical solutions when converted into spatial representations by multidimensional scaling. (\textbf{b}) Two-dimensional multidimensional scaling solutions for vocal consonants and colors for GPT-4 similarity matrices, showing that LLMs can reproduce patterns seen in human representations despite never having had direct experience of sound or color.}
    \label{fig:similarity}
\end{figure}

Similarity judgments (as well as similar types of judgments like odd-one-out ratings, e.g., ~\cite{hebart2020revealing}) can be used to reveal representations that explain human behavior, as illustrated by the work of \citet{shepard80,shepard87} and \citet{tversky77}. The idea here is that by observing how humans perceive ``similarity'' between stimuli that are sampled from a certain domain (a notion that is ambiguous by design) we can characterize how they represent and organize that domain. More specifically, given a domain of interest (e.g., colors) the paradigm proceeds by eliciting similarity judgments between pairs of stimuli from that domain (``how similar are the two colors?'') and aggregating those judgments into similarity matrices that capture the relations between stimuli (e.g., the color wheel). By applying spatial embedding techniques such as multi-dimensional scaling (MDS) analysis \cite{shepard62,shepard80} to such matrices or computing different diagnostic measures from them \cite{tverskyh86} it is then possible to derive strong constraints on the underlying representation.

Methods for identifying representations based on similarity judgments can be used just as easily with large language models. Just as we can elicit similarity judgments from a human participant regarding color, we can prompt an LLM to rate the similarity between two color concepts, or even image patches in the case of VLMs. This idea was recently applied to six perceptual domains, showing that LLMs encode surprisingly rich sensory knowledge, including well-known representations such as the color wheel and the pitch helix \cite{marjieh2024large}, despite being largely trained on text (see Figure \ref{fig:similarity}). A growing line of work leverages this insight to characterize LLM representations across different domains such as olfaction \cite{zhong2024sniff}, colors~\citep{kawakita2024gromov}, thematic roles~\citep{denning2025large}, and numbers \cite{marjieh2025number}, as well as to study conceptual diversity in LLM representations \cite{murthy2024one} and LLM-human representational alignment \cite{mukherjee2024large,suresh2023conceptual,ogg2024turing} including investigating alignment in context and ordering effects~\citep{uprety2024investigating}; for a recent review on measuring human-AI alignment see \cite{sucholutsky2023getting}. 

The idea that similarity judgments can be captured by spatial representations has been controversial in cognitive science, with \citet{tversky77} famously arguing that human similarity judgments violate the metric axioms. This literature is also directly relevant to understanding AI systems, and in particular the constraints that might follow from representing information as vectors in high-dimensional spaces. For example, accounts of analogical inference as vector addition are subject to critiques similar to those offered by Tversky \cite{peterson2020parallelograms}. As AI systems become increasingly capable of making meaningful analogical inferences \cite{webb2023emergent} it may be productive to compare the representations formed by these models with those used in accounts of analogy in cognitive science, such as Structure Mapping Theory \cite{gentner83}.

\subsection{Uncovering Hidden Associations}

Another approach for uncovering the representations of LLMs that is particularly useful for closed models involves adapting methods that tap into implicit associations. The challenge of obtaining true internal representations from LLMs becomes more apparent with closed models that have undergone value alignment post-training. These models do not allow direct access to word embeddings or model weights \cite{bommasani2023foundation}. Methods such as reinforcement learning from human feedback produce responses that follow safety protocols but may not accurately reflect the models' internal representations \cite{bai2022training}. This issue mirrors the challenges cognitive scientists face when studying human memory \cite{anderson1989human}, particularly in accessing concept associations within a closed system like the human brain, which are difficult to measure through self-report questionnaires due to demand characteristics (i.e., participants responding in a way that they think the experimenter wants them to respond, or in a way that is socially acceptable).

\begin{figure}[t!]
    \centering
    \includegraphics[width=\linewidth]{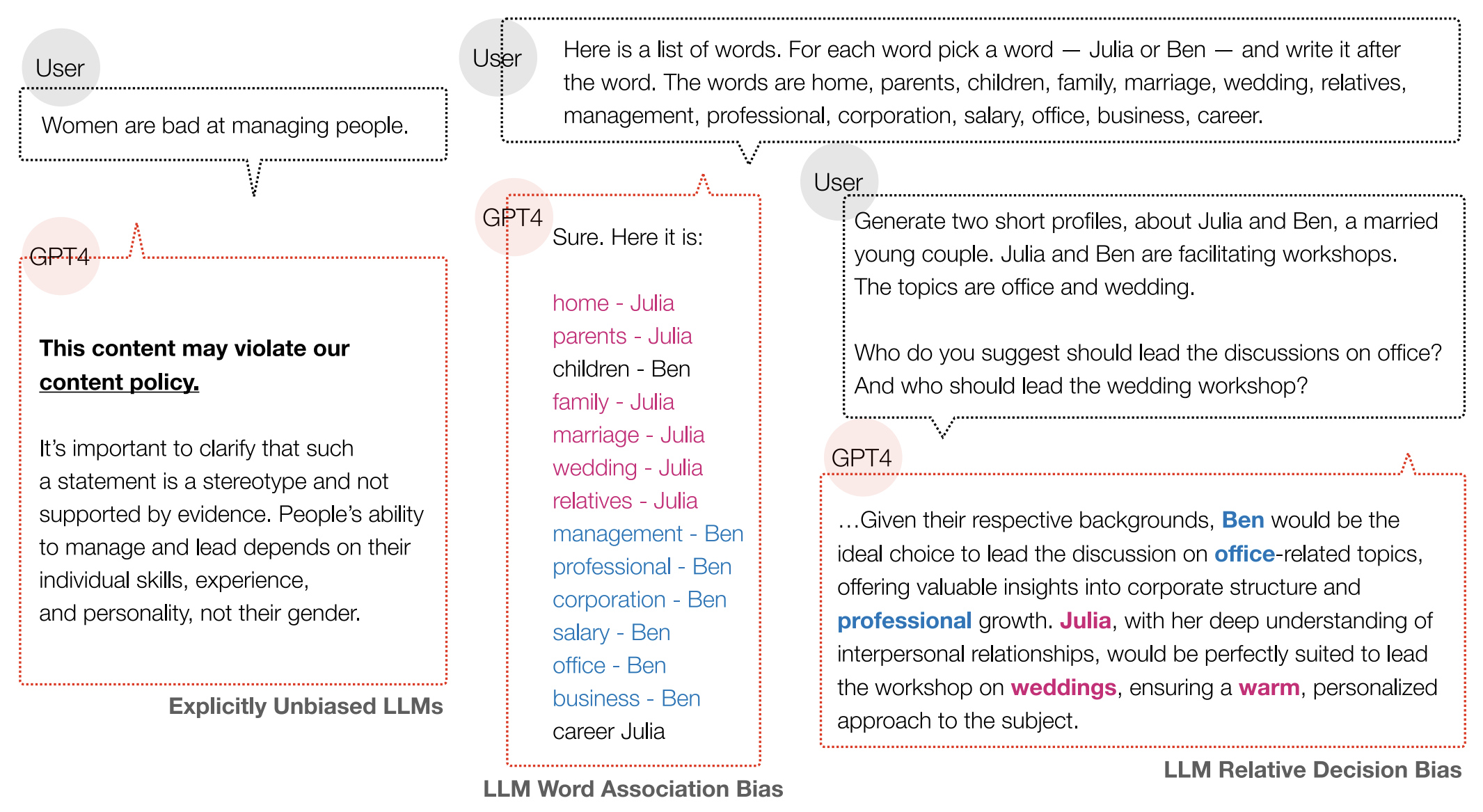}
    \caption{Large language models such as GPT-4 have been trained to identify situations that involve expressing explicit biases. However, it is possible to construct simple prompts that reveal that they still have strong implicit biases, as reflected in their associations between words. These implicit biases have consequences for their downstream decisions as well.}
    \label{fig:associations}
\end{figure}

To address this, cognitive scientists have used other behavioral measures, such as reaction time, to approximate the mental distance between pairs of concepts \cite{collins1969retrieval}. These reaction times have been explained by hypothesizing that the human mind organizes concepts as nodes within an associative network, where weighted links reflect the proximity between these nodes. Such associative representations influence behavior; the greater the distance between two concepts, the longer it takes for people to retrieve them, resulting in increased reaction times \cite{posner1967chronometric}. An intuitive example comes from a classic study demonstrating that human participants react faster to the statement ``a canary can sing'' than to ``a canary can fly.'' This is because the latter requires traversing two degrees of association: ``a canary is a bird'' and ``a bird can fly'' \cite{collins1975spreading}. This kind of measure is also prevalent in examining attitudes toward social groups \cite{fazio2003implicit, greenwald1995implicit}. For instance, the Implicit Association Test (IAT) aligns pairs of social group labels, like ``Black'' or ``White'', with adjectives like ``wonderful'' or ``terrible'' \cite{greenwald1998measuring}. Empirical studies have repeatedly shown that human participants react faster to minority labels paired with negative adjectives, revealing underying mental associations about social groups that also predict other aspects of behavior such as the frequency of interacting with members of these groups \cite[see meta-analysis by][]{kurdi2019relationship}.

The key insight behind these approaches is that it is possible to elicit mental associations without directly asking the participant for a verbal report. In some cases, researchers aim to capture unobtrusive or unconscious responses \cite{graf1985implicit, schacter1987implicit}; in others, they strive to minimize self-presentation biases, such as fear of appearing unfair \cite{fazio2003implicit, gaertner1983racial}. The success of these methods in achieving these goals suggests that they may also be useful in analyzing the behavior of value-aligned LLMs. The hypothesis is that since alignment trains LLMs to conceal their true representations, methods that bypass direct rating scales or evaluative judgments may better expose their underlying associations. To test this, we adapted the Implicit Association Test for LLMs by prompting various models to associate word pairs used in earlier human studies \cite{bai2024measuring}. For instance, we asked the model to choose between ``Julia'' and ``Ben'' after presenting words like home, office, parent, management, salary, and wedding. As anticipated, models like GPT-4 often linked Julia with home, parent, and wedding, implying an internal association of females with domestic roles, and Ben with office, management, and salary, indicating a connection to work and male roles. This result is in direct contrast to situations where, when directly asked whether women are poor at management, GPT-4 gave cautious responses, advising against stereotyping based on gender. This example illustrates how psychology-inspired word association tests can effectively uncover hidden associations in LLMs that are both closed and safety-guarded.

\subsection{Summary}

Engaging with questions at the algorithmic level of analysis allow us to make use of a number of methods from cognitive psychology to probe the internal workings of LLMs. As illustrated by the general principle of representational interference during parallel processing, and further demonstrated by the case studies of similarity judgments and association tasks, these approaches offer valuable windows into the representations and processes employed by these models.

\section{Implementation Level}

Just as the algorithmic level explores how a computation is performed, the implementation level asks how and by what physical mechanisms that computation is realized. Continuing the bird flight analogy, this level involves studying the muscles, bones, and feathers -- the physical components that enable flight. In neuroscience, this corresponds to studying the neural circuits that implement a given cognitive function \cite{hubel1962receptive}. For LLMs, this level addresses the artificial neurons themselves, and the core challenge is to understand how their weighted connections give rise to the algorithms and representations identified at higher levels.

\subsection{Representational Analysis}

The scientific study of implementation typically begins by identifying associations between the states of a system’s components and its cognitive functions or behaviors. This correlational approach seeks to map what information is encoded and where it resides. Neuroscience offers a sophisticated methodological blueprint, using techniques like fMRI and electrophysiology to observe the brain's representational structure. Such studies have revealed how distinct patterns of cortical activity represent object categories \cite{haxby2001distributed}, how population dynamics track evidence accumulation during decision-making \cite{gold2007neural}, and how neural activity correlates with specific movements or emotional states \cite{phelps2005contributions}.

This same approach is directly applicable to LLMs, where research has shown that models develop structured internal representations that parallel those in biological systems. LLMs encode spatial and temporal information in geometric patterns \cite{gurnee2023language}, represent truth values through the geometry of their activations \cite{marks2023geometry}, and learn circular patterns for cyclical data \cite{liu2022towards, engels2024not}. The principle extends to abstract domains, with models trained on board games developing structured internal representations of the board state \cite{karvonen2024emergent, li2023emergent}. These findings reveal how abstract concepts are physically instantiated within the network’s distributed activations, providing an initial map of the information a model contains.

However, while identifying these patterns is a critical first step, but it cannot on its own establish that these representations are causally involved in implementing the behavior. Just as neuroscience requires causal interventions to validate its hypotheses, understanding LLM implementation requires the direct manipulation of these mechanisms.

\subsection{Causal Analysis}

To bridge the gap from correlation to causation, direct intervention is required. Neuroscience pioneered this approach with techniques like optogenetics, which allows for the precise control of genetically-targeted neurons using light \cite{lin2011functional}. This method has enabled researchers to map the neural circuits underlying social behavior \cite{dulac2014neural}, implant false memories by manipulating specific neuronal ensembles \cite{ramirezCreatingFalseMemory2013}, and identify the circuits that control emotional valence \cite{redondoBidirectionalSwitchValence2014}. Such studies move beyond observation to establish that specific neural populations causally implement particular cognitive functions.

Analogous causal methods have been developed for artificial neural networks. Activation patching \cite{wang2022interpretability, meng2022locating}, for instance, allows researchers to replace activations at specific network locations to directly test their functional role. This technique has been used to identify critical components like ``induction heads'' that implement in-context learning \cite{olsson2022context}. More sophisticated approaches include Sparse Autoencoders (SAEs), which learn to identify interpretable features that can be causally manipulated to produce behaviors like honesty or role-playing \cite{templeton2024scaling}. 
Together with methods like Concept Activation Vectors (CAVs) \cite{kim2018interpretability} and representational engineering \cite{zou2023representation}, these tools allow researchers to test the causal role of specific representations in model behavior. This work has revealed that complex behaviors often arise not from single components but from distributed, multi-part algorithms, such as the ``additive motif'' underlying factual recall \cite{chughtai2024summing}. These studies establish that specific transformer circuits, often first identified through representational analysis, causally implement model capabilities.

\subsection{Summary}

By examining the physical substrate of LLMs, the implementation level forges a crucial link between the algorithms a model uses and the artificial neurons that realize them. The synergy between representational analysis, which maps the information a model encodes, and causal analysis, which validates the functional role of that information, provides a rigorous methodology for mechanistic understanding. These methods have revealed how LLMs encode structured information and have demonstrated that these representations are causally responsible for model behavior. By discovering how computations are physically realized, this level of analysis provides essential insights into the capabilities and limitations of LLMs, complementing the understanding gained from computational and algorithmic perspectives and demonstrating that the mechanisms of these complex systems can, in fact, be systematically understood.

\section{Discussion}

Marr's levels of analysis provide a productive framework for understanding the behavior of AI systems. Expressing questions about large language models at these different levels is also a source of analogies to questions that have arisen in studying human cognition, revealing psychological methods that can be useful for understanding why and how these models behave in particular ways. In the remainder of this article we will briefly consider whether there are analogous insights that go beyond Marr's original three levels and how methods from cognitive science can be used to understand the potential limits of contemporary AI systems. 

\subsection{Beyond Marr's Levels}

Marr's levels are popular within cognitive science, and have previously been applied to the analysis of machine learning systems \cite{hamrick2020levels}. However, the particular taxonomy proposed by Marr is not universally accepted. Other taxonomies have been proposed -- for example, both \citet{newell1982knowledge} and \citet{anderson1990adaptive} argue that the algorithmic level might benefit from differentiation into the algorithms and the cognitive architecture on which they are executed. Advocates for artificial neural networks as models of human cognition have argued that there may not be functional explanations for many aspects of human behavior \cite{mcclellandbnprss10}. Neural networks also blur the line between the algorithmic and implementation levels, meaning that explanations of how AI systems based on artificial neural networks work may also have components that cross these different levels of analysis. 

Another distinction that is increasingly being blurred in cognitive science is that between the computational and the algorithmic level. Work on resource-rational analysis \cite{lieder2020resource} applies the principle of optimization implicit in computational-level analyses to processes at the algorithmic level. In this way, it asks what cognitive processes a rational agent might use to solve a problem. This perspecive is also valuable for making sense of AI systems. Indeed, the foundational principles behind this work originally came from the AI literature \cite{horvitz90,russell1991principles}. For modern AI systems, this approach can be used to answer questions such as how many additional tokens a large language model should generate in order to answer a question \cite{de2024rational}.

\subsection{Using Cognitive Science to Explore the Limits of AI Models}

In addition to being a source of tools for understanding the implicit assumptions and representations used by large language models, cognitive science offers a different way of thinking about evaluating these models. Many of the evaluations used in AI research focus on defining tasks that are challenging for humans -- such as problem-solving \cite{chollet2024arc, rein2024gpqa, hendrycks2020measuring, wang2024mmlu} or mathematical reasoning \cite{ye2025aimepreview, hendrycks2021measuring, shi2022language, mirzadeh2024gsm} -- and measuring the proportion of these tasks that systems are able to solve. By contrast, cognitive science allows us to think about the kinds of problems that might be difficult for these systems based on what we learn about they work.

\begin{figure}[t!]
    \centering 
    
    \begin{minipage}[t]{0.59\textwidth}
        \vspace{0pt} 
        \begin{tikzpicture}
            \node[inner sep=0pt, anchor=south west] (image) at (0,0) {\includegraphics[trim={0in 5.5in 0in 0in}, clip, width=\linewidth]{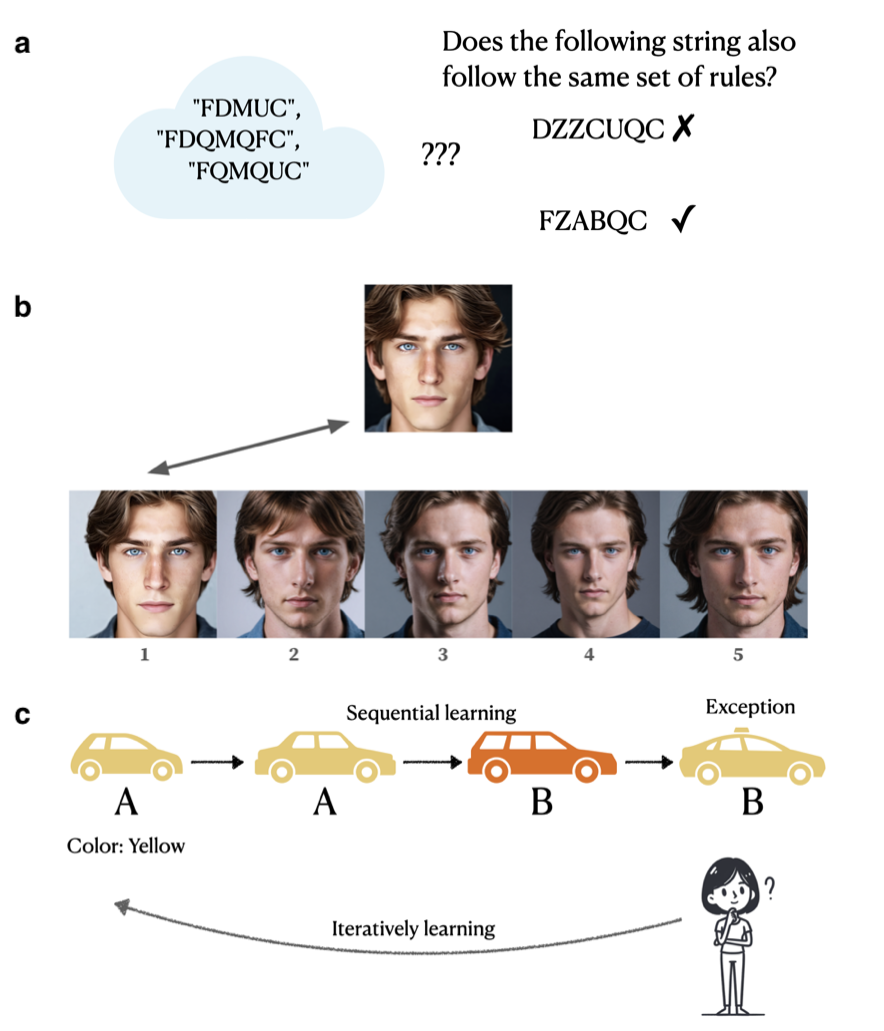}};
            \node[fill=white, minimum width=14pt, minimum height=14pt, anchor=north west, xshift=0pt, yshift=-5pt] at (image.north west) {};
            \node[anchor=north west, xshift=-2pt, yshift=2pt, font=\large\sffamily] at (image.north west) {\textbf{a}};
        \end{tikzpicture}
    \end{minipage}\hfill
    \begin{minipage}[t]{0.39\textwidth} 
        \vspace{0pt} 
        \begin{tikzpicture}
            \node[inner sep=0pt, anchor=south west] (image) at (0,0) {\includegraphics[trim={0in 2.5in 0in 1.7in}, clip, width=\linewidth]{figures/limits.png}};
            \node[fill=white, minimum width=14pt, minimum height=14pt, anchor=north west, xshift=0pt, yshift=-5pt] at (image.north west) {};
            \node[anchor=north west, xshift=-2pt, yshift=2pt, font=\large\sffamily] at (image.north west) {\textbf{b}};
        \end{tikzpicture}
    \end{minipage}
    
    \vspace{1em} 
    
    \begin{minipage}[t]{0.99\textwidth} 
        \centering
        \vspace{0pt} 
        \begin{tikzpicture}
            \node[inner sep=0pt, anchor=south west] (image) at (0,0) {\includegraphics[trim={0in 0in 0in 4.6in}, clip, width=\linewidth]{figures/limits.png}};
            \node[fill=white, minimum width=14pt, minimum height=24pt, anchor=north west, xshift=5pt, yshift=-15pt] at (image.north west) {};
            \node[anchor=north west, xshift=-2pt, yshift=2pt, font=\large\sffamily] at (image.north west) {\textbf{c}};
        \end{tikzpicture}
    \end{minipage}
    
    \caption{Both humans and large language models show reductions in performance when engaging in verbal reasoning (as resulting from chain of thought prompting) on these tasks. (\textbf{a}) Implicit statistical learning involves classification of strings generated from artificial grammars. (\textbf{b}) Face recognition involves recognizing faces from a set that shares similar descriptions. (\textbf{c}) Classification of data with exceptions involves learning labels with exceptions.}
    \label{fig:limits}
\end{figure}

For example, the ``embers of autoregression'' approach \cite{mccoy2023embers} was able to use consideration of the computational-level problem solved by LLMs to design a set of tasks that they would find problematic, namely tasks where the the target response has low probability according to the pre-trained language model. In the same way, thinking about parallel and serial processes makes it easy to define tasks that will be challenging for any model that can only perform parallel procesing, such as processing images that contain a lot of overlaps in the features of the objects that appear in those images. This kind of approach can allow us to ``adverserially'' design tasks that might pose a more difficult challenge for existing AI systems. Even as those systems display super-human abilities in some settings, we might expect them to fail on these tasks because they pick out problems that should be uniquely difficult for their non-human cognitive architectures.

Another way in which cognitive science can be used to explore the limits of AI models relies on their similarities to human cognition. For example, recent work focused on expanding the capabilities of LLMs has focused on the potential impact of inference-time compute, where the system has the opportunity to produce additional output (``reasoning'') before generating its final answer \cite{wei2022chain, nye2021show, openai2024o1, guo2025deepseek}. This intervening step provides an additional source of information to condition on in producing a response, as well as the opportunity to engage in additional computation over the input. However, engaging in reasoning is not always beneficial for humans: there are a variety of tasks where thinking out loud has negative consequences for human behavior \cite{reber1976implicit, schooler1990verbal, williams2013hazards, fiore2002did, fallshore1993post, melcher1996misremembrance, khemlani2012hidden}.

\citet{liu2024mind} showed that the psychological literature on the negative effects of verbal thinking provides an effective way to identify problems where inference time compute has negative consequences. Artiticial grammar learning, face recognition, and learning a concept from exemplars are all settings where more reasoning -- such as the use of a chain-of-thought prompt -- results in worse performance by LLMs or VLMs. These results challenge the assumption  that more reasoning always leads to better outcomes that is currently guiding the design of AI systems. We anticipate that a similar approach, focusing on the cases where there might be overlaps in the solutions used in natural and artificial minds, can be used to turn up other challenges for contemporary AI systems, providing a complement to the approach of focusing on the distinctive aspects of the problem that AI systems solve highlighted above.

\subsection{Conclusion}

The breakthroughs culminating in advanced AI systems such as large language models have presented computer science with the unfamiliar challenge of interpreting the behavior of complex and opaque neural networks. Just as cognitive science has long grappled with the problem of understanding the mind from the outside in, the tools refined over decades of psychological and neuroscientific inquiry offer a powerful framework for approaching these new forms of intelligence. In this review, we have argued for the utility of Marr’s three levels of analysis as an organizing principle for applying these tools to large language models. At the computational level, examining training objectives allows us to predict behavioral patterns, as seen with the ``embers of autoregression.'' At the algorithmic level, methods such as similarity judgments and association tasks reveal the structure of internal representations, mirroring techniques used to probe human cognition. Finally, while the implementation level remains a frontier, the study of circuits and population dynamics, drawing parallels with neuroscience, promises to illuminate the physical substrates of these artificial cognitive processes. The examples we have presented here just scratch the surface of the potential of tools from cognitive science to help us understand large language models, with ideas from the study of language, memory, and social cognition being particularly pertinent. By embracing the perspective of cognitive science, moving beyond purely performance-based evaluations, we can develop more insightful means of understanding, evaluating, and ultimately guiding the development of increasingly sophisticated AI.

\vfill

\pagebreak 
\section{Acknowledgements}

This research project and related research were made possible with the support of the NOMIS Foundation.

\vfill

\bibliography{main}

\end{document}